\documentclass{article}

\usepackage[final]{corl_2018}

\usepackage[utf8]{inputenc} % allow utf-8 input
\usepackage[T1]{fontenc}    % use 8-bit T1 fonts
\usepackage{hyperref}       % hyperlinks
\usepackage{url}            % simple URL typesetting
\usepackage{booktabs}       % professional-quality tables
\usepackage{nicefrac}       % compact symbols for 1/2, etc.
\usepackage{microtype}      % microtypography

\usepackage{amsmath,amsfonts,amssymb}
\usepackage{algorithm,algorithmicx}
\usepackage{algpseudocode}
\usepackage{graphicx}
\usepackage{caption}
\usepackage{subcaption}
\usepackage{mathtools}
\usepackage{tikz}
\usepackage{pgf}
\usetikzlibrary{arrows,automata}
\usepackage{multicol}

\newcommand{\abf}{\mathbf{a}}

\newcommand{\cbf}{\mathbf{c}}
\newcommand{\dbf}{\mathbf{d}}

\newcommand{\fbf}{\mathbf{f}}
\newcommand{\gbf}{\mathbf{g}}
\newcommand{\hbf}{\mathbf{h}}

\newcommand{\kbf}{\mathbf{k}}

\newcommand{\pbf}{\mathbf{p}}
\newcommand{\qbf}{\mathbf{q}}
\newcommand{\rbf}{\mathbf{r}}
\newcommand{\sbf}{\mathbf{s}}

\newcommand{\ubf}{\mathbf{u}}

\newcommand{\xbf}{\mathbf{x}}
\newcommand{\ybf}{\mathbf{y}}
\newcommand{\zbf}{\mathbf{z}}

\newcommand{\Abf}{\mathbf{A}}
\newcommand{\Bbf}{\mathbf{B}}
\newcommand{\Cbf}{\mathbf{C}}
\newcommand{\Dbf}{\mathbf{D}}
\newcommand{\Ebf}{\mathbf{E}}
\newcommand{\Fbf}{\mathbf{F}}

\newcommand{\Hbf}{\mathbf{H}}
\newcommand{\Ibf}{\mathbf{I}}

\newcommand{\Kbf}{\mathbf{K}}

\newcommand{\Mbf}{\mathbf{M}}

\newcommand{\Pbf}{\mathbf{P}}
\newcommand{\Qbf}{\mathbf{Q}}

\newcommand{\Sbf}{\mathbf{S}}

\newcommand{\Xbf}{\mathbf{X}}

\newcommand{\Zbf}{\mathbf{Z}}

\newcommand{\Acal}{\mathcal{A}}

\newcommand{\Ncal}{\mathcal{N}}
\newcommand{\Ocal}{\mathcal{O}}

\newcommand{\Scal}{\mathcal{S}}

\newcommand{\Xcal}{\mathcal{X}}

\newcommand{\alphabf}{\boldsymbol{\alpha}}

\newcommand{\Sigmabf}{\mathbf{\Sigma}}

\newcommand{\varepsilonbf}{\boldsymbol{\varepsilon}}

\newcommand{\zerobf}{\mathbf{0}}

\newcommand{\Hblk}{
	\left(
	\begin{array}{cc}
		\Hbf_{t-1} \ \zerobf \\
		\hline
		 \hbf^\top_{t}
	\end{array}
	\right)
}

\newcommand{\Htblk}{
	\left(
	\begin{array}{c|c}
		\Hbf^\top_{t-1} & \hbf_{t}\\
		\zerobf^\top
	\end{array}
	\right)
}

\DeclareMathOperator*{\argmax}{arg\,max}

\title{Recursive Sparse Pseudo-input Gaussian Process SARSA }

\author{
  John Martin and Brendan Englot\\
  Department of Mechanical Engineering\\
  Stevens Institute of Technology\\
  Hoboken, NJ 07030 \\
  \texttt{$\{$jmarti3,benglot$\}$@stevens.edu} 
}

\begin{document}
\maketitle

%
% ABSTRACT
%
\begin{abstract}
	The class of Gaussian Process (\textsc{gp}) methods for Temporal Difference learning has shown promise for data-efficient model-free Reinforcement Learning. In this paper, we consider a recent variant of the GP-SARSA algorithm, called Sparse Pseudo-input Gaussian Process SARSA (\textsc{spgp-sarsa}), and derive recursive formulas for its predictive moments. This extension promotes greater memory efficiency, since previous computations can be reused and, interestingly, it provides a technique for updating value estimates on a multiple timescales.    
\end{abstract}

\section{Introduction}
In the Reinforcement Learning (RL) paradigm, an agent interacts with an unknown environment by taking actions and collecting rewards. Throughout this process, the agent strives to maximize its total expected reward, or value $Q$. The optimal value can be recovered with Bellman's equation \cite{dp:1957:bellman:dynamic_programming}, whereby observed rewards are used to update estimates of $Q$ through the unbiased, recursive relation 
\begin{align*}
Q(\sbf,\abf) = r + \gamma \Ebf[Q(\Sbf,\Abf)]	.
\end{align*}
This applies as the agent undergoes a random transition from $(\sbf,\abf) \rightarrow (\Sbf,\Abf)$.

Bellman's equation motivates many methods for finding the optimal value. Among the most data-efficient are the class of Gaussian Process methods, which replace sample-intensive estimation schemes with a Bayesian non-parametric estimator, based on Gaussian Process regression \cite{rl:deisenroth_rasmussen-pilco,rl:2003:engel_mannor_meir:bayes_meets_bellman_the_gaussian_process_approach_to_temporal_difference_learning,rl:ghavamzadeh_engel_valko-bayesian_policy_gradient_and_actor_critic_algorithm,rl:2017:martin:heteroscedastic,rl:2005:engel_etal:learning_to_control_an_octopus_arm_with_gaussian_process_temporal_difference_methods,rl:2018:martin_etal:spgp_sarsa,rl:2004:rasmussen_kuss:gaussian_processes_in_reinforcement_learning} . These methods have shown to yield state-of-the-art empirical performance in their respective domains, such as model-based and model-free learning.

In this paper, we consider the class of Gaussian Process methods for model-free temporal difference learning \cite{rl:2003:engel_mannor_meir:bayes_meets_bellman_the_gaussian_process_approach_to_temporal_difference_learning,rl:2018:martin_etal:spgp_sarsa}. Specifically, we target the Sparse Pseudo-input Gaussian Process SARSA (\textsc{spgp-sarsa}) method \cite{rl:2018:martin_etal:spgp_sarsa} to improve its online viability with a procedure to perform recursive updates. By extending \textsc{spgp-sarsa} this way, its functionality then covers the full scope of benefits captured by its predecessor, \textsc{gp-sarsa} \cite{rl:2003:engel_mannor_meir:bayes_meets_bellman_the_gaussian_process_approach_to_temporal_difference_learning}; namely, previous results can be reused to compute matrix inverses.

\section{TD Value Estimation as GP Regression}\label{sec:gptd}
\textsc{td} algorithms recover the latent value function with data gathered in the standard \textsc{rl} fashion: at each step, the robot selects an action $\abf\in\Acal$ based on its current state $\sbf\in\Scal$, after which it transitions to the next state $\sbf'$ and collects a reward $R\sim p_r(\cdot|\sbf,\abf)$. The repeated interaction is described as a Markov Decision Process, $(\Scal,\Acal,p_r,p_s,\gamma)$, associated with the transition distribution $\sbf'\sim p_s(\cdot|\sbf,\abf)$, stationary policy $\abf\sim\pi(\cdot|\sbf)$, and discount factor $\gamma \in [0,1]$. As the name suggests, \textsc{td} algorithms update a running estimate of the value function to minimize its error difference from the Bellman estimate: $r + \gamma Q(\sbf',\abf') - Q(\sbf, \abf)$; $r$ being the observed reward. Once the estimate converges, an agent can select actions from the greedy policy $\pi$, such that $\Ebf_\pi[\Abf|\sbf] = \argmax_{\abf\in \Acal} Q(\sbf,\abf)$.

%In continuous settings, \textsc{td} estimation requires $Q$ to be approximated - typically with a parametric function. Finding parameters associated with the true value function can often require more data than marine robots can provide in a timely manner.    
The Gaussian Process Temporal Difference (\textsc{gptd}) framework improves upon the data efficiency of frequentist \textsc{td} estimation by departing from the contractive nature of Bellman's equation, in favor of a convergence driven by non-parametric Bayesian regression. The data model is based on the random return $Z(\xbf) = \sum_{t=0}^\infty \gamma^tR(\xbf_t)$, expressed as a sum of its mean, $Q(\xbf)$, and zero-mean residual, $\Delta Q(\xbf) = Z(\xbf) - Q(\xbf)$. Model inputs are state-action vectors $\xbf\in \Xcal=\Scal\times\Acal$, and value differences are used to describe the observation process:
\begin{align}\label{eq:trans_likelihood}
	R(\xbf) &= Q(\xbf) -\gamma Q(\xbf') + [\Delta Q(\xbf)-\gamma\Delta Q(\xbf')] = Q(\xbf) -\gamma Q(\xbf') + \varepsilon(\xbf,\xbf').
\end{align}
Moving forward, we assume that noise levels, $\varepsilon(\xbf,\xbf')$, are i.i.d random variables with constant parameters, $\varepsilon \sim \Ncal(0,\sigma^2)$. Under this assumption, transitions exhibit no serial correlation, and the \textsc{spgp-sarsa} model is valid.

%Notice how the noise can be state-dependent. We postpone the discussion of state-dependent noise to an extended version of the paper. However, see \cite{rl:2005:engel:algorithms_and_representations_for_reinforcement_learning} for a state-dependent derivation of standard \textsc{gp-sarsa}.

Given a time-indexed sequence of transitions $(\xbf_{t},R(\xbf_{t}),\xbf_{t+1})_{t=0}^{N-1}$, the \textsc{gp-sarsa} model stacks variables into vectors to obtain the complete data model: $\rbf = \Hbf\qbf(\xbf)+\varepsilonbf$, where 
\begin{align}\label{eq:gptd_model}
	\begin{pmatrix}
		R(\xbf_0)\\
		R(\xbf_1)\\
		\vdots\\
		R(\xbf_{N-1})
	\end{pmatrix} &= 
	\begin{pmatrix}
		1 & -\gamma & 0 & \cdots & 0\\
		0 & 1 & -\gamma & \cdots &0 \\
		\vdots & & & & \vdots\\
		0 & 0 &\cdots & 1 & -\gamma
	\end{pmatrix}
	\begin{pmatrix}
		Q(\xbf_0)\\
		Q(\xbf_1)\\
		\vdots\\
		Q(\xbf_N)
	\end{pmatrix} + 
	\begin{pmatrix}
		\varepsilon_0\\
		\varepsilon_1\\
		\vdots\\
		\varepsilon_N
	\end{pmatrix},
\end{align}
and $\qbf \sim \Ncal(\zerobf,\Kbf_{qq})$. Notice the commonality Equation \ref{eq:gptd_model} has with a standard \textsc{gp} likelihood model, $\ybf = \fbf(\xbf) + \varepsilonbf$. Both models assume the outputs, $\rbf \sim \ybf$, are noisy observations of a latent function, $\qbf \sim \fbf$. What distingushes \textsc{td} estimation is the presence of value correlations, imposed from Bellman's equation and encoded as temporal difference coefficients in $\Hbf$. Used for exact \textsc{gp} regression, Equation \ref{eq:gptd_model} leads to the \textsc{gp-sarsa} algorithm: a non-parametric Bayesian method for recovering latent values \cite{rl:2003:engel_mannor_meir:bayes_meets_bellman_the_gaussian_process_approach_to_temporal_difference_learning}. 

As a Bayesian method, \textsc{gp-sarsa} computes a predictive posterior over the latent values by conditioning on observed rewards. The corresponding mean and variance are used for policy evaluation:
\begin{align}
	v(\xbf_*) &= \kbf^\top_{r*}(\Kbf_{rr} + \sigma^2\Ibf)^{-1}\rbf, &
	s(\xbf_*) &= k(\xbf_*,\xbf_*) -\kbf^\top_{r*}(\Kbf_{rr} + \sigma^2\Ibf)^{-1}\kbf_{r*}.
	\label{eq:gptd_predmoms}
\end{align}
Here, $\Kbf_{qq}$ is the covariance matrix with elements  $[\Kbf_{qq}]_{ij} = k(\xbf_i,\xbf_j)$, $\Kbf_{rr} = \Hbf\Kbf_{qq}\Hbf^\top$, and $\kbf_{r*} = \Hbf\kbf_*$, where $[\kbf_*]_{i} = k(\xbf_i,\xbf_*)$. Subscripts denote dimensionality, e.g. $\Kbf_{qq} \in \mathbb{R}^{|\qbf|\times |\qbf|}$.

%
% An Online Algorithm
%
\section{Sparse Pseudo-input Gaussian Process Temporal Difference Learning} \label{sec:sgptd}
\begin{figure*}
	\centering
	\includegraphics[width=0.3\columnwidth]{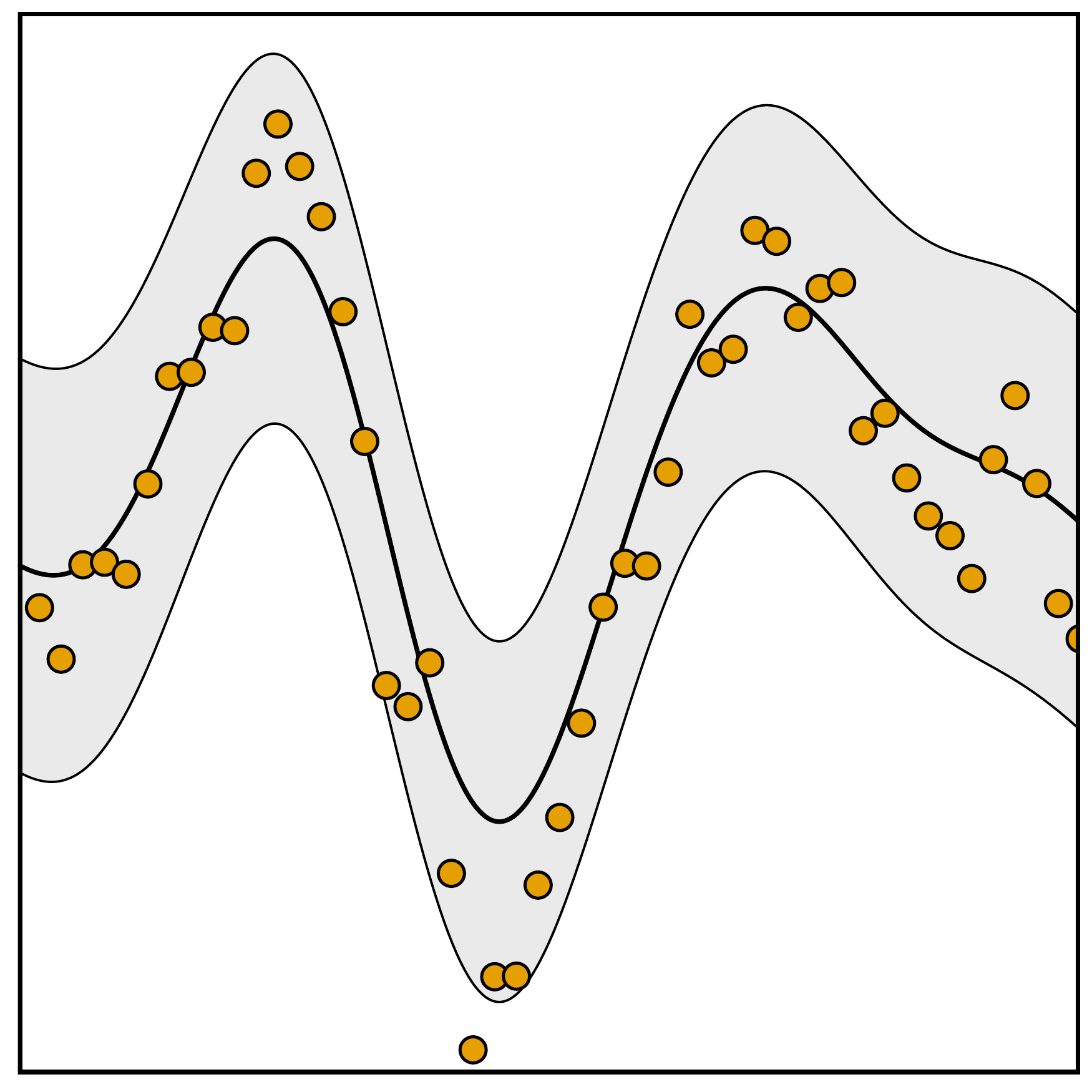}
	\includegraphics[width=0.3\columnwidth]{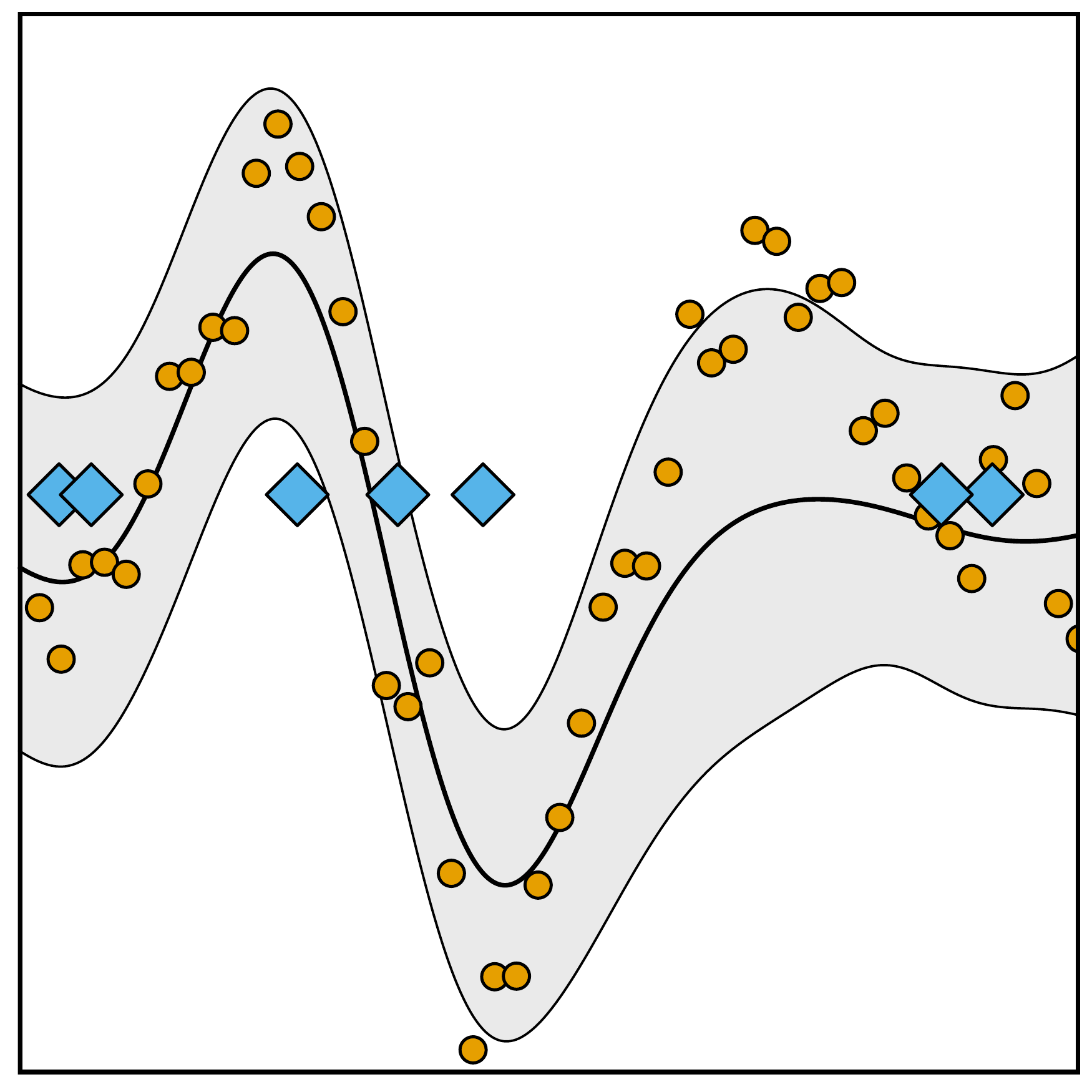}
	\includegraphics[width=0.3\columnwidth]{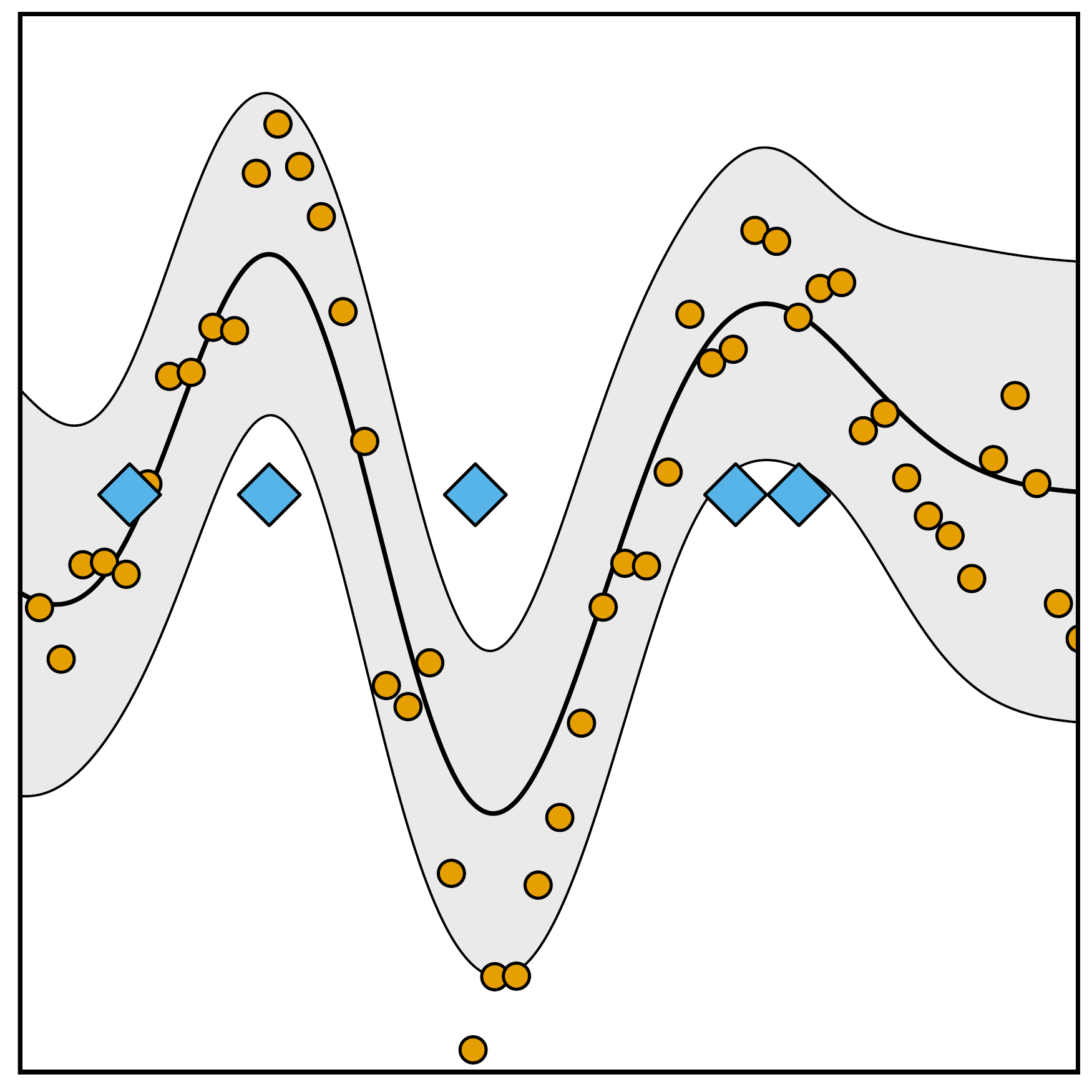}
	\caption{\textbf{Visualizing \textsc{gp}- and \textsc{spgp-sarsa} posteriors:} The exact \textsc{gp-sarsa} posterior (left) is supported with all the training data (dots). The \textsc{spgp-sarsa} posterior (center) uses randomly-initialized pseudo inputs (diamonds). After adjusting the pseudo inputs to maximize likelihood of training data, the posterior (right) is nearly identical to the exact model.}
\end{figure*}
The \textsc{gp-sarsa} method requires an expensive $N\times N$ matrix inversion, costing $\Ocal(N^3)$. To improve computational efficiency, \textsc{spgp-sarsa} algorithm applies the Sparse Pseudo-input Approximation \cite{gp:2006:snelson_ghahramani:sparse_gaussian_processes_using_pseudo_inputs}. Sparsity is induced in the standard data model (Equation \ref{eq:gptd_model}) by expanding the probability space with $M\ll N$ additional pseudo values, $\ubf$. The corresponding pseudo inputs, $\zbf\in\Zbf\subset \Xcal$, act as parameters on the support of the predictive posterior. These extra latent variables obey the same data model as $\qbf$, but are predetermined, and thus, exhibit no noise. By conditioning $\qbf$ upon $\ubf$ and $\Zbf$, the predictive probability space collapses such that all dense matrix inversions are of rank $M$. This algorithm is called Sparse Pseudo-input Gaussian Process SARSA (\textsc{spgp-sarsa}) \cite{rl:2018:martin_etal:spgp_sarsa}.

The \textsc{spgp-sarsa} predictive posterior is Gaussian, $\Ncal(\tilde{v}(\xbf),\tilde{p}(\xbf))$, with parameter functions
\begin{align}\label{eq:pred_moments}
	\tilde{v}(\xbf_*) &=\kbf_{u*}^\top\Mbf^{-1}\Kbf_{ur}(\Qbf + \sigma^2\Ibf)^{-1}\rbf &
	\tilde{p}(\xbf_*) &= k(\xbf_*,\xbf_*) - \kbf_{u*}^\top(\Kbf_{uu}^{-1} - \Mbf^{-1})\kbf_{u*}.
\end{align}

\section{Deriving a Recursive Algorithm}
The terms in Equation \ref{eq:pred_moments} that do not depend on the input effectively parameterize the posterior. We denote these parameters as
\begin{align}\label{eq:predictive_mean}
	\alphabf_t &= \Cbf_{kk}\Kbf_{kt}\Hbf_t^\top\Bbf_{tt}\rbf_{t-1}, &
	\Pbf_t &= \Abf_{kk} - \Cbf_{kk}.
\end{align}

Here we adopt a new notation that allows us to index updates associated with each input variable, $\xbf$ and $\zbf$. We drop the $v$ and $u$ designations on matrices in favor of $t$ and $k$, which respectively denote the update index of $\xbf$ and $\zbf$. A detailed breakdown of the notation is given below
\begin{itemize}
	\item{$t$: the unique index for the inputs $\xbf$.}
	\item{$k$: the unique index for the pseudo inputs $\zbf$}
	\item{$\kbf_{t-1}(\xbf_t) = (k(\xbf_1,\xbf_t),\cdots,k(\xbf_{t-1},\xbf_t))^\top$. Here, the array covers the range $1,\cdots,t-1$, and $\xbf_t$ is the common argument to all elements.}
	\item{$\Kbf_{k}$: the $k\times k$ covariance matrix of pseudo input evaluations $[\Kbf_k]_{ij}=k(\zbf_i,\zbf_j)$} 
	\item{$\Kbf_t$: the $t\times t$ covariance matrix of input evaluations $[\Kbf_t]_{ij}=k(\xbf_i,\xbf_j)$}
\end{itemize}

This new notation suggests that \textsc{spgp-sarsa} can be used on two timescales. There is the scale, $t$, associated with state transitions, and the scale $k$, associated with adding new pseudo inputs. Although we do not elaborate on when and how to apply multi-timescale updates, we believe this constitutes the subject of interesting future work.  

There are four distinct modalities in which \textsc{spgp-sarsa} can be updated:
\begin{enumerate}
	\item{\textit{Offline}: $\Zbf$ is fixed. $\Xbf$ is fixed.} 
	\item{\textit{Offline}: $\Zbf$ can vary. $\Xbf$ is fixed.} 
	\item{\textit{Online}: $\Zbf$ is fixed. $\Xbf$ can vary.} 
	\item{\textit{Online}: $\Zbf$ can vary. $\Xbf$ can vary.} 
\end{enumerate}
Here we consider the third and fourth cases, when both the transition training set and the pseudo set can grow. We decompose the predictive moments into partitioned matrices and apply the partitioned matrix inversion lemma to derive a recursive algorithm for their updates.
\subsection{Partitioned Matrix Inversion Lemma}
Let $\Kbf_t$ be a $t\times t$ symmetric positive definite matrix whose partition is
\begin{align}
	\Kbf_t = 
	\begin{pmatrix}
		\Kbf_{t-1} & \kbf_t\\
		\kbf_t^\top & k_{tt}
	\end{pmatrix}.
\end{align}
Define $s_t = k_{tt} - \kbf^\top\Kbf_{t-1}^{-1}\kbf_t$. Then the inverse is given by 
\begin{align}
	\Kbf_t^{-1} = 
	\begin{pmatrix}
		\Kbf_{t-1}^{-1} & \zerobf\\
		\zerobf & 0
	\end{pmatrix} + 
	\frac{1}{s_t}
	\begin{pmatrix}
		\Kbf_{t-1}^{-1}\kbf_t\\
		-1
	\end{pmatrix}
	\begin{pmatrix}
		\kbf_t^\top\Kbf_{t-1}^{-1} & -1
	\end{pmatrix}.
\end{align}

\subsection{Partitioning the Fundamental Matrices}
For deterministic transitions, the noise matrix is
\begin{align}
		\Sigmabf_t =
		\begin{pmatrix}
			\Sigmabf_{t-1} & \zerobf\\
			\zerobf & \sigma^2_{t-1}
		\end{pmatrix}.
\end{align}

The Bellman matrix partition is
\begin{align}
		\Hbf_t &= 
	\left(
	\begin{array}{cc}
		\Hbf_{t-1} \ \zerobf \\
		\hline
		 \hbf^\top_{t}
	\end{array}
	\right),
	& \hbf_t &= (0, \cdots, 1 , -\gamma)^\top.
\end{align}
Here, $\Hbf_t \in \mathbb{R}^{t-1\times t}$, with $\hbf_t \in \mathbb{R}^t$.

\begin{align}
		\Kbf_{tt} &=
		\begin{pmatrix}
			\Kbf_{t-1t-1} & \kbf_{t-1}(\xbf_t)\\
			\kbf_{t-1}^\top(\xbf_t) & k_{tt}
		\end{pmatrix}
\end{align}
Here, we define $\kbf_{t-1}(\xbf_t) = (k(\xbf_1,\xbf_t) , \cdots , k(\xbf_{t-1},\xbf_t))^\top$, $k_{tt}=k(\xbf_t , \xbf_t )$.

\begin{align}
	\Kbf_{kk} = 
	\begin{pmatrix}
		\Kbf_{k-1k-1} & \kbf_{k-1}(\zbf_k)\\
		\kbf_{k-1}^\top(\zbf_k) & k_{kk}
	\end{pmatrix}
\end{align}
Here we define $\kbf_{k-1}(\zbf_t) = (k(\zbf_1,\zbf_t) , \cdots , k(\zbf_{k-1},\zbf_t))^\top$, $k_{kk}=k(\zbf_k , \zbf_k )$.

\begin{align}
	\Kbf_{tk} &=
	\begin{pmatrix}
		\Kbf_{t-1k-1} & \kbf_{t-1}(\zbf_k)\\
		\kbf_{k-1}^\top(\xbf_t) & k_{tk}
	\end{pmatrix}			
\end{align}
This partition is special, because it can be updated in three ways. We may include a new input $\xbf_t$, a pseudo input $\zbf_k$, or both. Here, $\xbf$ includes the rows with index $t$, and the pseudo inputs, $\zbf$, include the columns with index $k$. We define $\kbf_{k-1}(\xbf_t) = (k(\zbf_1,\xbf_t) , \cdots , k(\zbf_{k-1},\xbf_t))^\top$, $k_{tk}=k(\xbf_t , \zbf_k )$, and $\kbf_{t-1}(\zbf_k) = (k(\xbf_1,\zbf_k) , \cdots , k(\xbf_{t-1},\zbf_k))^\top$. When $\xbf_t$ is new, we must compute $\kbf_{k-1}^\top(\xbf_t)$ and $k_{tk}$. Similarly, when $\zbf_k$ is new, we must compute $\kbf_{t-1}(\zbf_k)$ and $k_{tk}$. When both variables are new, the only element we may reuse is $\Kbf_{t-1k-1}$. 

This concludes our analysis of the fundamental matrices. Next we turn our attention to the compound matrices, which are products of those described above.

\subsection{Partitioning Compound Matrices}

\begin{align}
		\Hbf_t\Kbf_{tk}
		&= \Hblk
		\begin{pmatrix}
			\Kbf_{t-1k-1} & \kbf_{t-1}(\zbf_k)\\
			\kbf_{k-1}^\top(\xbf_t) & k_{tk}
		\end{pmatrix}\nonumber\\
		&=
		\begin{pmatrix}
			\Hbf_{t-1}\Kbf_{t-1k-1} & \Hbf_{t-1}\kbf_{t-1}(\zbf_k)\\
			(\kbf_{k-1}(\xbf_{t-1}) - \gamma\kbf_{k-1}(\xbf_t))^\top & k(\xbf_{t-1},\zbf_k) - \gamma k(\xbf_t,\zbf_k)
		\end{pmatrix}\nonumber\\
		&=
		\begin{pmatrix}
			\Hbf_{t-1}\Kbf_{t-1k-1} & \Hbf_{t-1}\kbf_{t-1}(\zbf_k)\\
			\Delta\kbf_{k-1}^\top(\xbf_{t}) & \Delta k_k(\xbf_t)
		\end{pmatrix}
\end{align}
Here we define $\Delta\kbf_{k-1}^\top(\xbf_{t}) = \kbf_{k-1}(\xbf_{t-1}) - \gamma\kbf_{k-1}(\xbf_t)$, and $\Delta k_k(\xbf_t)= k(\zbf_k,\xbf_{t-1}) - \gamma k(\zbf_k,\xbf_t)$.
	
%\begin{align}
%		\Kbf_{kt}\Hbf_t^\top
%		&=
%		\begin{pmatrix}
%			\Kbf_{k-1t-1} & \kbf_{k-1}(\xbf_t)\\
%			\kbf_{t-1}^\top(\zbf_k) & k_{kt}
%		\end{pmatrix}\Htblk \nonumber\\
%		&=
%		\begin{pmatrix}
%			\Kbf_{k-1t-1}\Hbf_{t-1}^\top & \kbf_{k-1}(\xbf_{t-1}) - \gamma\kbf_{k-1}(\xbf_t)\\
%			(\Hbf_{t-1}\kbf_{t-1}(\zbf_k))^\top & k(\xbf_{t-1},\zbf_k) - \gamma k(\xbf_t,\zbf_k)
%		\end{pmatrix}\nonumber,\\
%		&=
%		\begin{pmatrix}
%			\Kbf_{k-1t-1}\Hbf_{t-1}^\top & \Delta\kbf_{k-1}(\xbf_{t})\\
%			(\Hbf_{t-1}\kbf_{t-1}(\zbf_k))^\top & \Delta k_k(\xbf_t)
%		\end{pmatrix}
%\end{align}

\begin{align}
	\Hbf_t\Kbf_{tt}\Hbf_t^\top &= 
	\Hblk \begin{pmatrix}
		\Kbf_{t-1t-1} & \kbf_{t-1}(\xbf_t)\\
		 \kbf_{t-1}^\top(\xbf_t) & k_{tt}
	\end{pmatrix}
	\Htblk,\nonumber\\
	&=
	\Hblk
	\begin{pmatrix}
		\Kbf_{t-1t-1}\Hbf_{t-1}^\top & \kbf_{t-1}(\xbf_{t-1})-\gamma \kbf_{t-1}(\xbf_{t})\\
		\kbf^\top_{t-1}(\xbf_t)\Hbf^\top_{t-1} & k(\xbf_{t-1},\xbf_t) - \gamma k(\xbf_t,\xbf_t)
	\end{pmatrix},\nonumber\\
	&=
	\begin{pmatrix}
		\Hbf_{t-1}\Kbf_{t-1t-1}\Hbf_{t-1}^\top & \Hbf_{t-1}(\kbf_{t-1}(\xbf_{t-1})-\gamma \kbf_{t-1}(\xbf_{t})) \\
		(\kbf_{t-1}(\xbf_{t-1})-\gamma \kbf_{t-1}(\xbf_{t}))^\top\Hbf_{t-1}^\top & k_{t-1t-1} -2\gamma k_{t-1t} + \gamma^2 k_{tt}
	\end{pmatrix},\nonumber\\
	&= 
	\begin{pmatrix}
		\Hbf_{t-1}\Kbf_{t-1t-1}\Hbf_{t-1}^\top & \Hbf_{t-1}\Delta\kbf_{t-1}\\
		 (\Hbf_{t-1}\Delta\kbf_{t-1})^\top & \Delta^2 k_{t}		
	\end{pmatrix}
\end{align}
Here we define $\Delta\kbf_{t-1}(\xbf_t) = \kbf_{t-1}(\xbf_{t-1})-\gamma \kbf_{t-1}(\xbf_{t})$. We also use $\Delta^2 k_{t}$ to denote the arguments distributed according to a binomial: 
\begin{align*}
	\Delta^2 k_{t} &= k\circ [(\xbf_{t-1},\cdot) - \gamma(\xbf_{t},\cdot)][(\cdot,\xbf_{t-1}) - \gamma(\cdot,\xbf_t)] \\
	&=
	k\circ[(\xbf_{t-1},\xbf_{t-1}) - \gamma(\xbf_{t-1},\xbf_t) - \gamma(\xbf_t,\xbf_{t-1}) +\gamma^2(\xbf_t,\xbf_t)],\\
	&=k(\xbf_{t-1},\xbf_{t-1}) - \gamma k(\xbf_{t-1},\xbf_t) - \gamma k(\xbf_t,\xbf_{t-1}) +\gamma^2 k(\xbf_t,\xbf_t),\\
	&=k(\xbf_{t-1},\xbf_{t-1}) - 2\gamma k(\xbf_{t-1},\xbf_t) +\gamma^2 k(\xbf_t,\xbf_t).
\end{align*}

\section{Computing Inverse Matrix Partitions}
Three matrices must be inverted. They are:
\begin{itemize}
	\item{$\Abf_{kk} = \Kbf_{kk}^{-1}$}
	\item{$\Bbf_{tt} = [\text{diag}(\Hbf_t\Kbf_{tt}\Hbf_t^\top - \Hbf_t\Kbf_{tk}\Kbf_{kk}^{-1}\Kbf_{kt}\Hbf_t^\top) + \Sigmabf_t]^{-1}$}
	\item{$\Cbf_{kk} = [\Kbf_{kk} + \Kbf_{kt}\Hbf_t^\top\Bbf_{tt}\Hbf_t\Kbf_{tk}]^{-1}$} 
\end{itemize}

\subsection{Inverting $\Kbf_{kk}$}
By the matrix inversion lemma, we have
\begin{align}
	\Abf_{kk} &= 
	\begin{pmatrix}
		\Abf_{k-1k-1} & \abf_{k-1}(\zbf_k)\\
		\abf_{k-1}^\top(\zbf_k) & a_k
	\end{pmatrix} =
	\begin{pmatrix}
		\Kbf_{k-1k-1}^{-1} + s_k\gbf_{k-1}(\zbf_k)\gbf^\top_{k-1}(\zbf_k) & -s_k\gbf_{k-1}(\zbf_k)\\
		-s_k\gbf_{k-1}^\top(\zbf_k) & s_k
	\end{pmatrix}.
\end{align}
Here, we have defined:
\begin{align}
	\gbf_{k-1}(\zbf_k) &= \Kbf_{k-1k-1}^{-1}\kbf_{k-1}(\zbf_k),\\
	1/s_k &= k_{kk} - \kbf_{k-1}^\top(\zbf_k)\Kbf_{k-1k-1}^{-1}\kbf_{k-1}(\zbf_k).	
\end{align}

\subsection{Computing $\Bbf_{tt}$}
To start, we compute the composite matrix $\Dbf_{tt} = \Hbf_t\Kbf_{tk}\Abf_{kk}\Kbf_{kt}\Hbf_t^\top$
\begin{align*}
	\Dbf_{tt} &= 
	\begin{pmatrix}
		\Hbf_{t-1}\Kbf_{t-1k-1} & \Hbf_{t-1}\kbf_{t-1}(\zbf_k)\\
		\Delta\kbf_{k-1}^\top(\xbf_{t}) & \Delta k_k(\xbf_t)
	\end{pmatrix}
	\begin{pmatrix}
		\Abf_{k-1k-1} & \abf_{k-1}(\zbf_k)\\
		\abf_{k-1}^\top(\zbf_k) & a_k
	\end{pmatrix}		
	\begin{pmatrix}
		\Kbf_{k-1t-1}\Hbf_{t-1}^\top & \Delta\kbf_{k-1}(\xbf_{t})\\
		(\Hbf_{t-1}\kbf_{t-1}(\zbf_k))^\top & \Delta k_k(\xbf_t)
	\end{pmatrix},\\
	&=
	\begin{pmatrix}
		\Hbf_{t-1}\Kbf_{t-1k-1} & \Hbf_{t-1}\kbf_{t-1}(\zbf_k)\\
		\Delta\kbf_{k-1}^\top(\xbf_{t}) & \Delta k_k(\xbf_t)
	\end{pmatrix} \\
	&\cdot
	\begin{pmatrix}
		\Abf_{k-1k-1}\Kbf_{k-1t-1}\Hbf_{t-1}^\top + \abf_{k-1}(\zbf_k)(\Hbf_{t-1}\kbf_{t-1}(\zbf_k))^\top & \Abf_{k-1k-1}\Delta\kbf_{k-1}(\xbf_{t})+\abf_{k-1}(\zbf_k)\Delta k_k(\xbf_t)\\
		\abf_{k-1}^\top(\zbf_k)\Kbf_{k-1t-1}\Hbf_{t-1}^\top + a_k(\Hbf_{t-1}\kbf_{t-1}(\zbf_k))^\top & \abf_{k-1}^\top(\zbf_k)\Delta\kbf_{k-1}(\xbf_{t}) +  a_k\Delta k_k(\xbf_t)
	\end{pmatrix},\\
	&= 
	\begin{pmatrix}
		\Hbf_{t-1}\Delta\Dbf_{t-1t-1}\Hbf_{t-1}^\top & \Hbf_{t-1}\Delta \dbf_{t-1}(\xbf_t,\zbf_k) \\
		(\Hbf_{t-1}\Delta \dbf_{t-1}(\xbf_t,\zbf_k))^\top  & \Delta d_t
	\end{pmatrix},
\end{align*}
where we have defined the following elements: 
\begin{align*}
	\Delta\Dbf_{t-1t-1} &= \Kbf_{t-1k-1}\Abf_{k-1k-1}\Kbf_{k-1t-1} + \widetilde{\Dbf}_{t-1t-1},\\
	\widetilde{\Dbf}_{t-1t-1} &= 2\Kbf_{t-1k-1}\abf_{k-1}(\zbf_k)\kbf_{t-1}^\top(\zbf_k)+ a_k\kbf_{t-1}(\zbf_k)\kbf_{t-1}^\top(\zbf_k)\\
	\Delta \dbf_{t-1}(\xbf_t,\zbf_k) &= \Kbf_{t-1k-1}[\Abf_{k-1k-1}\Delta\kbf_{k-1}(\xbf_{t})+\abf_{k-1}(\zbf_k)\Delta k_k(\xbf_t)] \\
	&+ \kbf_{t-1}(\zbf_k)[\abf_{k-1}^\top(\zbf_k)\Delta\kbf_{k-1}(\xbf_{t}) +  a_k\Delta k_k(\xbf_t)],\\
	\Delta d_{t} &=\Delta\kbf_{k-1}^\top(\xbf_{t})[\Abf_{k-1k-1}\Delta\kbf_{k-1}(\xbf_{t})+\abf_{k-1}(\zbf_k)\Delta k_k(\xbf_t)] \\
	&+ \Delta k_k(\xbf_t)[\abf_{k-1}^\top(\zbf_k)\Delta\kbf_{k-1}(\xbf_{t}) +  a_k\Delta k_k(\xbf_t)].
\end{align*}

Now we add all the terms and take the diagonal:
\begin{align*}
	\Bbf_{tt}^{-1} &= 
	\begin{pmatrix}
		\Hbf_{t-1}\Kbf_{t-1t-1}\Hbf_{t-1}^\top & \Hbf_{t-1}\Delta\kbf_{t-1}(\xbf_t)\\
		 (\Hbf_{t-1}\Delta\kbf_{t-1}(\xbf_t))^\top & \Delta k_t		
	\end{pmatrix}\\
	&
	-
	\begin{pmatrix}
		\Hbf_{t-1}\Delta\Dbf_{t-1t-1}\Hbf_{t-1}^\top & \Hbf_{t-1}\Delta \dbf_{t-1}(\xbf_t,\zbf_k) \\
		(\Hbf_{t-1}\Delta \dbf_{t-1}(\xbf_t,\zbf_k))^\top  & \Delta d_t
	\end{pmatrix} +
	\begin{pmatrix}
			\Sigmabf_{t-1} & \zerobf\\
			\zerobf & \sigma^2_{t-1}
	\end{pmatrix},\\
	&=
	\begin{pmatrix}
		\Hbf_{t-1}(\Kbf_{t-1t-1}-\Delta{\Dbf}_{t-1t-1})\Hbf_{t-1}^\top + \Sigmabf_{t-1}& \Hbf_{t-1}(\Delta\kbf_{t-1}(\xbf_t)-\Delta\dbf_{t-1}(\xbf_t,\zbf_k))\\
		 (\Hbf_{t-1}(\Delta\kbf_{t-1}(\xbf_t)-\Delta\dbf_{t-1}(\xbf_t,\zbf_k))^\top & \Delta k_t - \Delta d_t + \sigma^2_{t-1}
	\end{pmatrix}.
\end{align*}
We define $1/b_j = \Delta k_j - \Delta d_j + \sigma^2_{j}$ for $j=1,\cdots , t$. The inverse is simply:
\begin{align}
	\Bbf_{tt} &=
	\begin{pmatrix}
		\Bbf_{t-1t-1} & \zerobf\\
		\zerobf &  b_{t}
	\end{pmatrix}.
\end{align}

\subsection{Computing $\Cbf_{kk}$}
First we compute $\Fbf_{kk} = \Kbf_{kt}\Hbf_t^\top\Bbf_{tt}\Hbf_t\Kbf_{tk}$.
\begin{align}
	\Fbf_{kk} &=
	\begin{pmatrix}
			\Kbf_{k-1t-1}\Hbf_{t-1}^\top & \Delta\kbf_{k-1}(\xbf_{t})\\
			(\Hbf_{t-1}\kbf_{t-1}(\zbf_k))^\top & \Delta k_k(\xbf_t)
		\end{pmatrix}
	\begin{pmatrix}
		\Bbf_{t-1t-1} & \zerobf\\
		\zerobf &  b_{t}
	\end{pmatrix}
	\begin{pmatrix}
		\Hbf_{t-1}\Kbf_{t-1k-1} & \Hbf_{t-1}\kbf_{t-1}(\zbf_k)\\
		\Delta\kbf_{k-1}^\top(\xbf_{t}) & \Delta k_k(\xbf_t)
	\end{pmatrix},\nonumber \\
	&= 
	\begin{pmatrix}
		\Kbf_{k-1t-1}\Hbf_{t-1}^\top & \Delta\kbf_{k-1}(\xbf_{t})\\
		(\Hbf_{t-1}\kbf_{t-1}(\zbf_k))^\top & \Delta k_k(\xbf_t)
	\end{pmatrix}
	\begin{pmatrix}
		\Bbf_{t-1t-1}\Hbf_{t-1}\Kbf_{t-1k-1} & \Bbf_{t-1t-1}\Hbf_{t-1}\kbf_{t-1}(\zbf_k)\\
		b_t\Delta\kbf_{k-1}^\top(\xbf_{t}) & b_t\Delta k_k(\xbf_t)
	\end{pmatrix},\nonumber \\
	&=
	\begin{pmatrix}
		\Fbf_{k-1k-1} &  \fbf_{k-1}\\
		 \fbf_{k-1}^\top &  f_{k}
	\end{pmatrix} + b_t
	\begin{pmatrix}
		\Delta\kbf_{k-1}(\xbf_{t}) \\
		\Delta k_k(\xbf_t) 
	\end{pmatrix}
	\begin{pmatrix}
		\Delta\kbf_{k-1}(\xbf_{t}) ,\Delta k_k(\xbf_t) 
	\end{pmatrix},\\
	&=
	\begin{pmatrix}
		\Delta\Fbf_{k-1k-1} &  \Delta\fbf_{k-1}\\
		 \Delta\fbf_{k-1}^\top &  \Delta f_{k}
	\end{pmatrix}
\end{align}
Here we define: 
\begin{align}
	\fbf_{k-1} &= \Kbf_{k-1t-1}\Hbf_{t-1}^\top\Bbf_{t-1t-1}\Hbf_{t-1}\kbf_{t-1}(\zbf_k) ,\\
	f_k &= \kbf_{t-1}^\top(\zbf_k)\Hbf_{t-1}^\top\Bbf_{t-1t-1}\Hbf_{t-1}\kbf_{t-1}(\zbf_k).
\end{align}
Adding the terms produces the partitioned $\Cbf_{kk}=\Kbf_{kk} + \Fbf_{kk}$ matrix:
\begin{align}
	\Cbf_{kk}^{-1} &= 
	\begin{pmatrix}
		\Kbf_{k-1k-1} + \Delta\Fbf_{k-1k-1} & \kbf_{k-1}(\zbf_k) + \Delta\fbf_{k-1}\\
		(\kbf_{k-1}(\zbf_k) + \Delta\fbf_{k-1})^\top & k_{kk} + \Delta f_{k}
	\end{pmatrix}.
\end{align}
By the matrix inversion lemma, we have:
\begin{align}
	\Cbf_{kk} &= 
	\begin{pmatrix}
		\Cbf_{k-1k-1} + w_k\tilde{\cbf}_{k-1}\tilde{\cbf}_{k-1}^\top & -w_k\tilde{\cbf}_{k-1}\\
		-w_k\tilde{\cbf}_{k-1}^\top & w_k
	\end{pmatrix},\\
	\tilde{\cbf}_{k-1} &= \Cbf_{k-1k-1}(\kbf_{k-1}(\zbf_k) + \Delta\fbf_{k-1}),\\
	1/w_k &= k_{kk} + \Delta f_{k} - \tilde{\cbf}_{k-1}^\top\Cbf_{k-1k-1}\tilde{\cbf}_{k-1}.
\end{align}

\subsection{Recursive Parameters}
We may now derive recursions for the parameters $\alphabf_t=\Cbf_{kk}\Kbf_{kt}\Hbf_t^\top\Bbf_{tt}\rbf_{t-1}$ and $\Pbf_{kk}$:
\begin{align}
	\alphabf_{t} &= 
	\begin{pmatrix}
		\Cbf_{k-1k-1} + w_k\tilde{\cbf}_{k-1}\tilde{\cbf}_{k-1}^\top & -w_k\tilde{\cbf}_{k-1}\\
		-w_k\tilde{\cbf}_{k-1}^\top & w_k
	\end{pmatrix}
	\begin{pmatrix}
		\Kbf_{k-1t-1}\Hbf_{t-1}^\top\Bbf_{t-1t-1} & b_t\Delta\kbf_{k-1}(\xbf_{t})\\
		(\Hbf_{t-1}\kbf_{t-1}(\zbf_k))^\top\Bbf_{t-1t-1} & b_t\Delta k_k(\xbf_t)
	\end{pmatrix}
	\begin{pmatrix}
		\rbf_{t-2}\\
		r_{t-1}
	\end{pmatrix},\nonumber\\
	&= 
	\begin{pmatrix}
		\alphabf_{t-1}\\
		0
	\end{pmatrix} +
	\begin{pmatrix}
		\tilde{\alphabf}_{t-1}\\
		\tilde{\alpha}_{t-1}
	\end{pmatrix}.
\end{align}
We have defined:
\begin{align*}
	\tilde{\alphabf}_{t-1}&= w_k\tilde{\cbf}_{k-1}[\tilde{\cbf}_{k-1}^\top\Kbf_{k-1t-1} - \kbf_{t-1}^\top(\zbf_k)]\Hbf_{t-1}^\top\Bbf_{t-1t-1}\rbf_{t-2}\\
	&+b_t[(\Cbf_{k-1k-1} + w_k\tilde{\cbf}_{k-1}\tilde{\cbf}_{k-1}^\top)\Delta\kbf_{k-1}(\xbf_{t}) -w_k\tilde{\cbf}_{k-1}\Delta k_k(\xbf_t)]r_{t-1},\\
	\tilde{\alpha}_t&=-w_k[\tilde{\cbf}_{k-1}^\top\Kbf_{k-1t-1} - \kbf_{t-1}^\top(\zbf_k)]\Hbf_{t-1}^\top\Bbf_{t-1t-1}\rbf_{t-2}\\
	& -w_kb_t[\tilde{\cbf}_{k-1}^\top\Delta\kbf_{k-1}(\xbf_{t}) - \Delta k_k(\xbf_t)]r_{t-1}.
\end{align*}
For the last parameter, we have:
\begin{align}
	\Pbf_{kk} &= \Abf_{kk} - \Cbf_{kk},\nonumber\\
	&=
	\begin{pmatrix}
		\Kbf_{k-1k-1}^{-1} + s_k\gbf_{k-1}(\zbf_k)\gbf^\top_{k-1}(\zbf_k) & -s_k\gbf_{k-1}(\zbf_k)\\
		-s_k\gbf_{k-1}^\top(\zbf_k) & s_k
	\end{pmatrix}
	-\begin{pmatrix}
		\Cbf_{k-1k-1} + w_k\tilde{\cbf}_{k-1}\tilde{\cbf}_{k-1}^\top & -w_k\tilde{\cbf}_{k-1}\\
		-w_k\tilde{\cbf}_{k-1}^\top & w_k
	\end{pmatrix},\nonumber\\
	&=
	\begin{pmatrix}
		\Pbf_{k-1k-1} + s_k\gbf_{k-1}(\zbf_k)\gbf^\top_{k-1}(\zbf_k) - w_k\tilde{\cbf}_{k-1}\tilde{\cbf}_{k-1}^\top & 
		w_k\tilde{\cbf}_{k-1} -s_k\gbf_{k-1}(\zbf_k)\\
		(w_k\tilde{\cbf}_{k-1} -s_k\gbf_{k-1}(\zbf_k))^\top & s_k - w_k
	\end{pmatrix},\nonumber\\
	&=
	\begin{pmatrix}
		\Pbf_{k-1k-1} & \zerobf\\
		\zerobf & 0
	\end{pmatrix}+
	\begin{pmatrix}
		 \widetilde{\Pbf}_{k-1k-1} & \tilde{\pbf}_{k-1}
		\\
		\tilde{\pbf}_{k-1}^\top & \tilde{p}_k
	\end{pmatrix},
\end{align}
where we have defined:
\begin{align*}
	\widetilde{\Pbf}_{k-1k-1} &= s_k\gbf_{k-1}(\zbf_k)\gbf^\top_{k-1}(\zbf_k) - w_k\tilde{\cbf}_{k-1}\tilde{\cbf}_{k-1}^\top,\\
	\tilde{\pbf}_{k-1} &=w_k\tilde{\cbf}_{k-1} -s_k\gbf_{k-1}(\zbf_k),\\
	\tilde{p}_k &= s_k - w_k.
\end{align*}

On-line updates are possible by unrolling the recursion, starting from basic posterior parameters.

\section{Conclusion}
In this paper we derived formulas for updating the \textsc{spgp-sarsa} algorithm recursively. This allows previous computations to be reused and promotes greater memory efficiency than computing matrix inverses from scratch at each iteration. Promising future work will explore the best practices for adding pseudo inputs and performing experiments to quantify the benefits of the recursive approach.

%%
%% BIBLIOGRAPHY
%%
\section{References}
\bibliographystyle{plain}
\bibliography{ref}  % .bib

\end{document}